\documentclass[conference]{IEEEtran}
\IEEEoverridecommandlockouts
\usepackage{cite}
\usepackage{amsmath,amssymb,amsfonts}
\usepackage{algorithmic}
\usepackage{graphicx}
\usepackage{textcomp}
\usepackage{xcolor}
\usepackage{booktabs}
\usepackage{hyperref}
\usepackage{multirow}
\usepackage{subfigure}
\usepackage{pifont}
\usepackage{bbding}

\def\BibTeX{{\rm B\kern-.05em{\sc i\kern-.025em b}\kern-.08em
    T\kern-.1667em\lower.7ex\hbox{E}\kern-.125emX}}
\begin{document}

\title{Parallel Spiking Unit for Efficient Training of Spiking Neural Networks}



\author{\IEEEauthorblockN{Yang Li$^{1,2,3}$, Yinqian Sun$^{1,2}$ Xiang He$^{1,2,3}$,  Yiting Dong$^{1,2,4}$,
Dongcheng Zhao$^{1,2}$, Yi Zeng$^{1,2,3,4,5,*}$}
\IEEEauthorblockA{\textit{$^{1}$Brain-inspired Cognitive Intelligence Lab, Institute of Automation, Chinese Academy of Sciences}\\
\textit{$^{2}$Center for Long-term Artificial Intelligence} \\
\textit{$^{3}$School of Artificial Intelligence, University of Chinese Academy of Sciences}\\
\textit{$^{4}$School of Future Technology, University of Chinese Academy of Sciences}\\
\textit{$^{5}$Key Laboratory of Brain Cognition and Brain-inspired Intelligence Technology, Chinese Academy of Sciences}\\
\textit{$^{*}$Corresonding author}}
\textit{\{liyang2019,sunyinqian2018,hexiang2021,dongyiting2020,zhaodongcheng2016,yi.zeng\}@ia.ac.cn}}



\maketitle

\begin{abstract}

Efficient parallel computing has become a pivotal element in advancing artificial intelligence. Yet, the deployment of Spiking Neural Networks (SNNs) in this domain is hampered by their inherent sequential computational dependency. This constraint arises from the need for each time step's processing to rely on the preceding step's outcomes, significantly impeding the adaptability of SNN models to massively parallel computing environments. Addressing this challenge, our paper introduces the innovative Parallel Spiking Unit (PSU) and its two derivatives, the Input-aware PSU (IPSU) and Reset-aware PSU (RPSU). These variants skillfully decouple the leaky integration and firing mechanisms in spiking neurons while probabilistically managing the reset process. By preserving the fundamental computational attributes of the spiking neuron model, our approach enables the concurrent computation of all membrane potential instances within the SNN, facilitating parallel spike output generation and substantially enhancing computational efficiency. Comprehensive testing across various datasets, including static and sequential images, Dynamic Vision Sensor (DVS) data, and speech datasets, demonstrates that the PSU and its variants not only significantly boost performance and simulation speed but also augment the energy efficiency of SNNs through enhanced sparsity in neural activity. These advancements underscore the potential of our method in revolutionizing SNN deployment for high-performance parallel computing applications.

\end{abstract}


\section{Introduction}


In the quest for more efficient neural network models, Spiking Neural Networks (SNNs), recognized as the third-generation artificial neural networks\cite{maass1997networks}, exhibit exceptional promise in developing innovative, intelligent computing paradigms. The unique appeal of SNNs lies in their biological plausibility and low-energy consumption characteristics\cite{roy2019towards,rathi2023exploring}, positioning them as an ideal model for emulating brain-like information processing. Unlike conventional artificial neural networks (ANNs), SNNs achieve an advanced simplification of complex neuronal dynamics. This is accomplished by replicating the sparse spiking activity of neurons, offering a computation method that is more efficient and conserves energy\cite{zhou2022spikformer}. Moreover, the intrinsic event-driven nature of SNNs means that computations are triggered only in response to significant changes in input signals. This significantly minimizes the usage of computational resources, paving the way for the creation of low-power yet high-performance artificial intelligence systems\cite{pei2019towards}.

By employing strategies such as converting trained ANNs into SNNs\cite{diehl2016conversion,li2022bsnn,he2023msat}, applying direct training with surrogate gradients to counter the non-differentiable nature of spikes\cite{wu2018spatio,shrestha2018slayer}, or incorporating biologically plausible mechanisms to guide or aid in SNN training\cite{zeng2017improving,zhao2022spiking,han2023enhancing}, SNNs have shown remarkable proficiency in handling complex networks and tasks. This has led to a progressive narrowing of the performance gap between SNNs and ANNs\cite{zhou2024spikformer,kim2020spiking,li2022spike}. Nonetheless, SNNs bear a computational resemblance to recurrent neural networks (RNNs), necessitating sequential information processing. Consequently, the training and inference phases of SNNs face limitations in parallelization in massively parallel computing environments, posing challenges to their efficient training, inference, and application.
Contemporary research has predominantly focused on the sophisticated refinement of spiking neuron models, particularly the widely utilized Leaky Integrate-and-Fire (LIF) neurons, aiming to significantly enhance the performance of computational models in the field. This includes adjusting specific neuronal attributes in a learnable manner, such as leakage coefficients\cite{fang2021incorporating} and thresholds\cite{ding2022biologically}. SPSN\cite{yarga2023accelerating} aims to parallelize the computation in spiking neurons through randomizing neuron outputs, while PSNs\cite{fang2023parallel} bypass the reset process and link information across individual time steps using weight parameters. Despite these advancements, current parallelized SNN models face limitations. They struggle to function effectively in complex tasks or exhibit increased firing rates, diminishing SNNs' inherent low-energy computational benefits.

\begin{figure*}[t]
    \centering
    \includegraphics[scale=0.38]{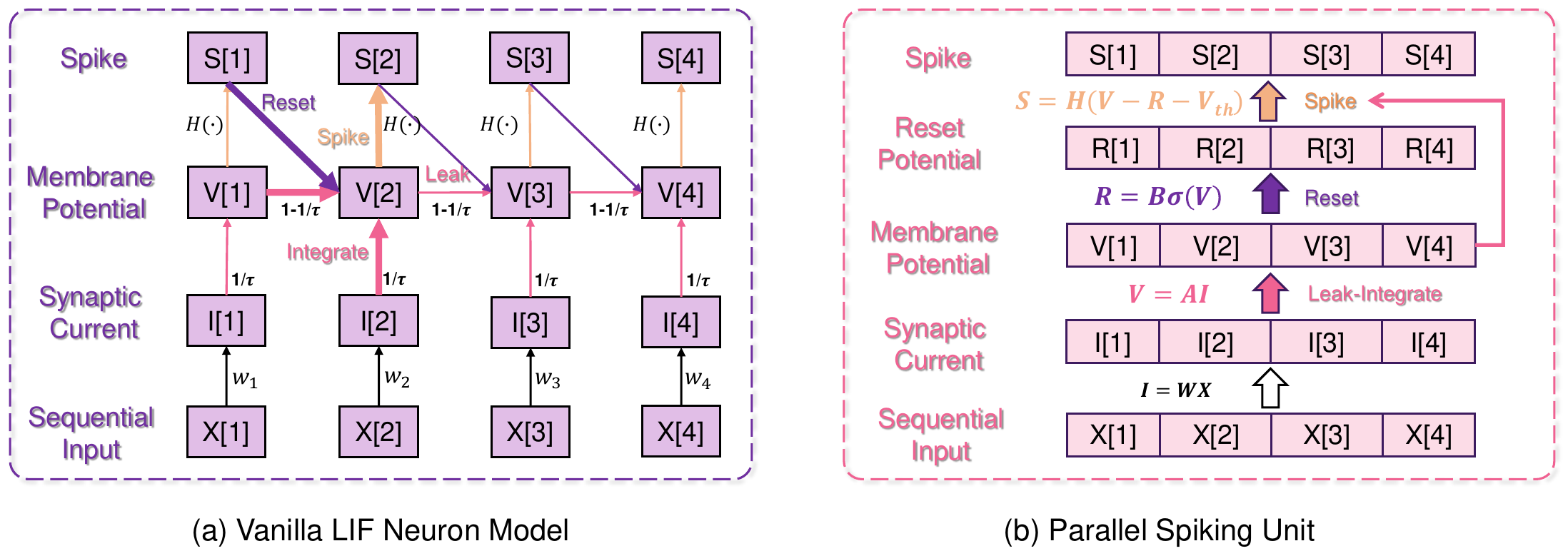}
    \caption{The computational process of the vanilla LIF neuron model and the parallel spiking units. Due to the Reset process, LIF neurons cannot be computed in parallel. In contrast, parallel spiking units decouple the Reset process from the Leak-Integrate process by estimating the required spike outputs, allowing for the parallel computation of outputs at all times. In the diagram, $H$ represents the Heaviside function, and $\sigma$ denotes the Sigmoid function.}
    \label{intro}
\end{figure*}

In this paper, we propose an innovative methodology, the Parallel Spiking Unit (PSU), along with its advanced variants: the Input-aware PSU (IPSU) and the Reset-aware PSU (RPSU), each characterized by a unique set of learnable parameters. Illustrated in Fig. \ref{intro}, conventional spiking neurons in a typical SNN perform a sequence of operations, leaky integration, spiking, and resetting, where the state at each time step is influenced by the preceding step's outcome. The reset mechanism, in particular, poses significant challenges for simultaneous spike computation across all time steps. Through theoretical examination, we have successfully decoupled the leak-integration function of the LIF neurons from their spiking behavior and reorganized the reset mechanism in probabilistic terms. This groundbreaking approach enables concurrent calculation of membrane potentials at each time step, based on input data, while maintaining essential computational properties of spiking neurons. Theoretically, PSUs enhance parallel processing capabilities while significantly lowering firing rates, augmenting their sparsity benefits, and accelerating simulation speed.
Furthermore, we introduce the IPSU and RPSU, which incorporate a causal mask into the leak-integration and reset mechanisms. The computation of inputs or spike predictions at each time step is informed by data from preceding steps, enhancing the learning of long-term dependencies and offering a more precise reset process estimation. Our extensive experimental evaluations span four diverse dataset categories: static images, sequential images, Dynamic Vision Sensor (DVS) data, and speech datasets. Compared to traditional LIF neurons, our novel PSU models and their variants demonstrate reduced spiking activity and quicker simulation speeds across various benchmarks, surpassing other spiking neuron models in performance metrics.

\section{Related Work}

\paragraph{Training methods for SNNs} Training SNNs introduces distinct challenges due to their binary activation functions, which undermine the efficiency of conventional backpropagation methods prevalent in ANNs. Investigating biologically inspired techniques, Diehl et al.\cite{diehl2015unsupervised} employ Spike-Timing-Dependent Plasticity (STDP) \cite{markram2011history} for the unsupervised training of a two-layer SNN, securing significant achievements on the MNIST benchmark. Similarly, Zhang et al.\cite{zhang2021self} explore self-backpropagation to enhance the backward flow of information to preceding layers. Nonetheless, SNNs generally underperform compared to ANNs on more intricate tasks.
To mitigate this discrepancy, converting pre-trained ANNs to structurally akin SNNs has emerged as a viable strategy, preserving the spike-based computational paradigm while attaining performance levels comparable to those of ANNs. In this context, Li et al.\cite{li2022efficient} introduce a Burst mechanism aimed at spiking neurons to diminish information loss. SpikingYOLO\cite{kim2020spiking} represents a notable leap in applying SNNs to object detection tasks. Furthermore, Bu et al.\cite{bu2023optimal} propose the pre-conversion quantization of ANN activation values, effectively reducing the inference latency of SNNs to a mere 4-8 time steps.
However, these conversion methodologies often encounter setbacks, including increased latency and reduced performance, prompting an escalated interest in surrogate gradients to address the non-differentiability issue inherent in spike functions. The introduction of Spatiotemporal Backpropagation (STBP)\cite{wu2018spatio} marks a pioneering step in this direction, employing surrogate gradients in the direct training of SNNs. Following this initiative, various studies\cite{suetake2022s2,li2021differentiable,herranz2022surrogate,kheradpisheh2022spiking} have further optimized SNN performance through surrogate gradients, achieving competitive results with ANNs in a notably short span.

\paragraph{Optimization of Spiking Neurons} Prior studies have predominantly concentrated on augmenting the information-processing capabilities of neurons to boost task performance. In the realm of conversion enhancements, Wang et al.\cite{wang2022signed} introduce the Triplet-spike neuron model, which is notable for emitting negative spikes to facilitate information correction. Concurrently, Li et al.\cite{li2022spike} pioneer the concept of Inter-spike Interval, a technique aimed at optimizing the adjustment of inactive neuronal spikes. In the context of direct training, the adaptation of parameters in LIF neurons, the LIAF model\cite{wu2021liaf} represents a significant advancement by employing only the output spikes for its reset mechanism while concurrently transmitting real-valued signals to subsequent layers, thereby improving the efficiency of information transfer. Furthermore, additional efforts\cite{ponghiran2022spiking,lotfi2020long} have ventured into integrating more sophisticated computational processes into the dynamics of spiking neurons, with a specific focus on mastering long-term dependencies.

\paragraph{Accelerating the Training Process} In the domain of sequence processing, addressing the slow training of RNNs due to sequential dependencies leads to the adoption of CNNs. CNNs facilitate parallel computation by processing information across various time instances through convolution. The Transformer model\cite{vaswani2017attention} further advances this by employing self-attention to model interactions between tokens at different times, enhancing the capture of long-term dependencies.
Adapting these insights to SNNs, SLTT\cite{meng2023towards} demonstrates that temporal optimization is independent of overall efficiency, incorporating SNNs' spatiotemporal dynamics only during forward propagation and focusing on spatial optimization during backpropagation. The SPSN model\cite{yarga2023accelerating} innovates further by making the spiking process probabilistic, allowing for the simultaneous input of temporal information and enabling parallel spike computation, albeit with a higher firing rate.
Building on SPSN, our approach fine-tunes spike output estimation while maintaining the reset process, leading to sparser computation and improved performance, enhancing the efficiency of neuronal models in SNNs.

\section{Methodology}

\subsection{Vanilla LIF Neuron}

The LIF neuron model, renowned for its ubiquity in neural simulations, strikes an optimal balance between computational tractability and preserving intricate neuronal dynamics. This model equips SNNs with the capability to process information encapsulating both spatial and temporal dimensions effectively. Information processing within a neuron is typically segmented into three distinct phases: the integration and leakage of membrane potentials, the spiking mechanism, and the reset process.

\begin{align}
    \label{LI}
    V[t] = V[t-1] + (-V[t-1] + I[t]) / \tau
\end{align}
\begin{align}
    \label{Spike}
    S[t] = \Theta (V[t]-V_{th})
\end{align}
\begin{align}
    V[t] = \begin{cases}V[t](1-S[t]), \quad \text{hard reset}\\ 
        V[t] - V_{th}S[t], \quad \text{soft reset} \end{cases}
\end{align}

Here, $V[t]$ is the membrane potential of the neuron, which adjusts itself with the synaptic current $I[t]$. When the neuron's membrane potential exceeds the threshold $V_{th}$, it fires a spike. Here, $\Theta$ is the Heaviside function. 
Two forms of reset are considered: hard reset and soft reset\cite{han2020rmp}, differing in whether the membrane potential is cleared after spike firing to further receive subsequent information.


\subsection{Parallel Spiking Unit}

Initially, we assume no spikes are fired by the neurons. Referring to Eq. \ref{LI}, the membrane potential is derived through matrix operations. Given that $V[0] = 0$, the initial step lacks a leakage process, leading to $V[1] = \frac{I[1]}{\tau}$. We dissect the information integrated at each step, assessing the leakage from previously integrated information up to the current moment. The information at step $t$ is $\frac{I[t]}{\tau}$, with the prior step's information $\frac{I[t-1]}{\tau}$ attenuating to $(1-\frac{1}{\tau})\frac{I[t-1]}{\tau}$ at step $t$, and so forth. Consequently, the membrane potential of N neurons at any step is a cumulative function of the synaptic currents $[I[1], I[2], \ldots, I[t']], t' \leq t$, facilitating the following formulation for membrane potential calculation:
\begin{align}
    V = AI, \quad V\in \mathbb{R}^{T \times N}, A\in \mathbb{R}^{T \times T}, I\in \mathbb{R}^{T \times N}
\end{align}

Considering $T=4$ as an example, it is evident that $V[1] = \frac{I[1]}{\tau}$ and $V[2] = V[1] + \left(-V[1]+I[2]\right) / \tau = \frac{1}{1-\tau}\frac{I[1]}{\tau}+\frac{I[2]}{\tau}$. From this, we can straightforwardly derive the expression for matrix $A$ about the hyperparameter $\tau$:
\begin{align}
    A = \begin{bmatrix} 
        \frac{1}{\tau} & 0 & 0 & 0 \\ 
        \frac{\tau-1}{\tau^2} & \frac{1}{\tau} & 0 & 0 \\ 
        \frac{(\tau-1)^2}{\tau^3} & \frac{\tau-1}{\tau^2} & \frac{1}{\tau} & 0 \\ 
        \frac{(\tau-1)^3}{\tau^4} & \frac{(\tau-1)^2}{\tau^3}& \frac{\tau-1}{\tau^2} & \frac{1}{\tau}\end{bmatrix}
\end{align}

If a neuron remains inactive except for the final time step, the spike emitting at each step can be determined using $\Theta(V-V_{th})$. While SNNs are characterized by sparsity, with some neurons fitting this assumption, the membrane potential at later steps often exceeds the post-reset calculation. Hence, relying solely on matrix $A$ might overestimate the firing rate compared to actual LIF neurons. To address this, SPSN introduces a probabilistic spike firing mechanism, estimating firing probabilities from parallel-computed membrane potentials and employing a Bernoulli process for spiking. Although this method alleviates the issue of inflated firing rates, it also indirectly highlights the importance of incorporating the reset process during the parallelization of the LIF model.

\begin{figure*}[t]
    \centering
    \includegraphics[scale=0.38]{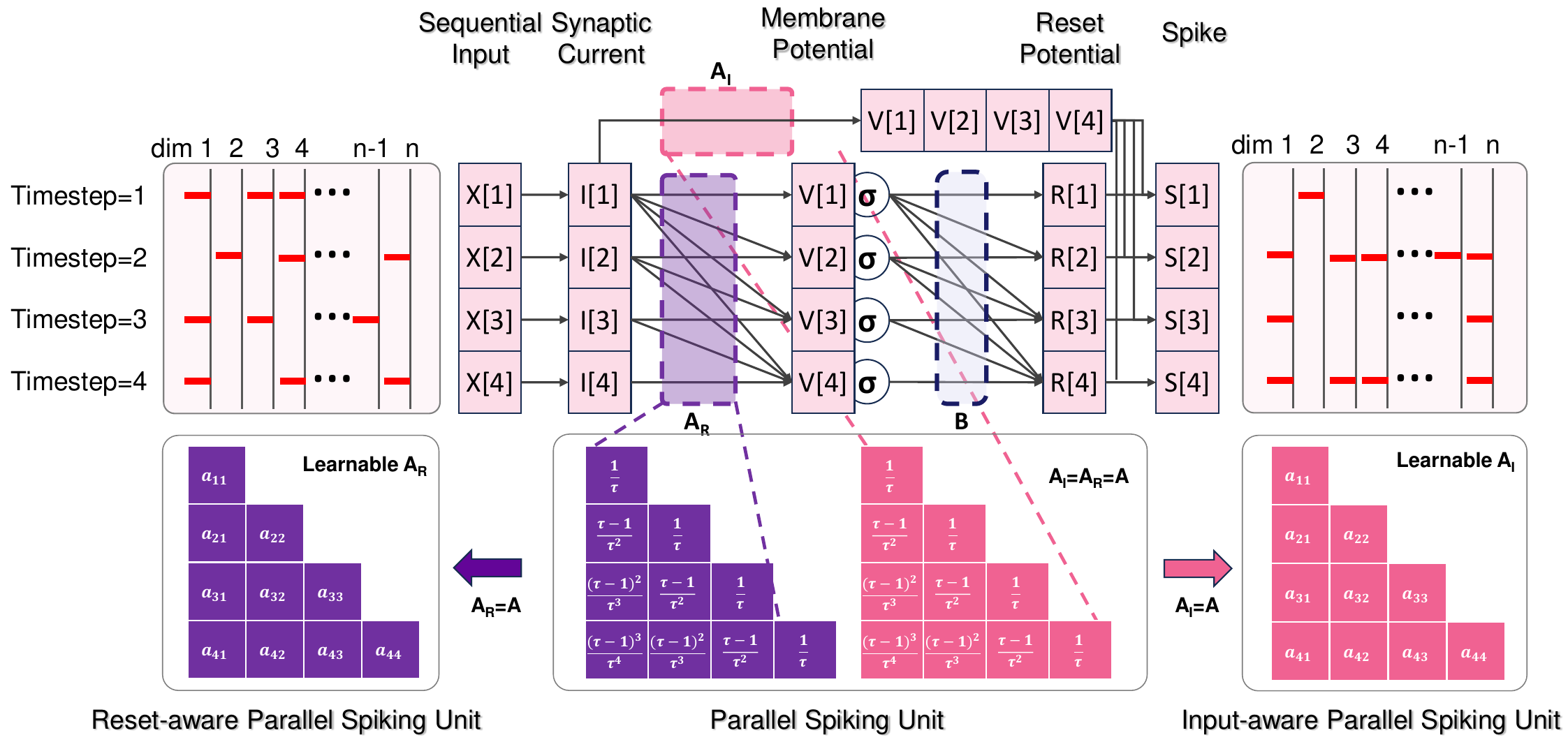}
    \caption{Comparison of PSU, IPSU, and RPSU. They both use causal masking techniques to mask the effect of information from future time steps on the current moment. RPSU learns parametrically the matrix $A_R$ involved in the reset part of the process from the point of view of better spike estimation. In contrast, IPSU learns parametrically the matrix $A_I$ responsible for the leakage integration from the point of view of better leakage integration. Both intend to preserve the reset process's features during parallelization and better adapt to the data through parametric learning.}
    \label{method}
\end{figure*}

The reset process in the LIF model diminishes the membrane potential's strength, essentially causing some of the historically integrated information to be forgotten. This reset mechanism depends on the neuron's spike activity at each time step. Focusing on the soft reset scenario, the membrane potential of a neuron that fires a spike at time $t$ will be adjusted by $-(1-\frac{1}{\tau}) V_{th}S[t]$ at time $t+1$. The leakage coefficient $\tau$ ensures that this loss of information is further modified to $-(1-\frac{1}{\tau})^2 V_{th}S[t]$ at time $t+2$, and this pattern continues over time. Consequently, the information loss $[R[1], R[2], \ldots, R[T]]$ induced by the reset process at each time step can be computed in parallel:
\begin{align}
    R = BS, \quad R \in \mathbb{R}^{T\times N}, B \in \mathbb{R}^{T\times T}, S \in \mathbb{R}^{T\times N}
\end{align}

Considering $T=4$ as an example, it's clear that $R[1]=0$, indicating no loss at the initial time step. At $T=2$, the loss $R[2]$ becomes $(1-\frac{1}{\tau}) V_{th} S[1]$, reflecting the impact of the first spike. Moving to $T=3$, the loss $R[3]$ accumulates to $(1-\frac{1}{\tau})^2 V_{th} S[1] + (1-\frac{1}{\tau}) V_{th} S[2]$, accounting for the decayed impact of the first spike and the immediate effect of the second. Hence, utilizing the hyperparameter $\tau$, we can formulate the expression for matrix $B$:
\begin{align}
    B = V_{th}\begin{bmatrix} 
        0& 0 & 0 & 0 \\ 
        1-\frac{1}{\tau} & 0 & 0 & 0 \\ 
        (1-\frac{1}{\tau})^2 & 1-\frac{1}{\tau} & 0 & 0 \\ 
        (1-\frac{1}{\tau})^3 & (1-\frac{1}{\tau})^2 & 1-\frac{1}{\tau} & 0\end{bmatrix}
\end{align}
Thus, before deciding whether a spike event occurs at each time step, the membrane potential of the neuron can be represented as
\begin{align}
    V = AI - BS
\end{align}
Substituting into Eq. \ref{Spike}, we can obtain: 
\begin{align}
    S = \Theta(AI - BS - V_{th})
\end{align}

The presence of the nonlinear function $\Theta$ complicates the expansion of the formula to derive an analytical expression for $S$, rendering the parallel computation of LIF neurons challenging. This analysis assumes the soft reset approach; however, the hard reset method encounters similar obstacles. Unlike the soft reset, which subtracts a constant value, the hard reset removes all inputs $[I[1], I[2], \ldots, I[t']], t' \leq t$ preceding $S[t]$. This necessitates incorporating element-wise multiplication between $I$ and $S$, further complicating the derivation of a formal expression.

In the discussion above, we segment the spiking process into \textit{Leak-integrate}, \textit{Spiking}, and \textit{Reset} phases, illustrating that the inability to parallelize the LIF model primarily stems from the challenge of \textit{Reset} phase depending on the \textit{Spiking} phase outcomes, which is exactly what we need. Therefore, we propose an estimation strategy for the \textit{Reset} phase, substituting the deterministic spike activity $S$ in the equation with an estimated spike activity $S' = \sigma(AI)$. Here, the Sigmoid function $\sigma(x) = \frac{1}{1+e^{-x}}$ is employed for probabilistic estimation, leading to the introduction of the parallel spiking unit (PSU), with the computational process outlined as follows:
\begin{align}
    \label{aibs}
    S =  \Theta(AI - BS' - V_{th}) =  \Theta(AI - B\sigma(AI) - V_{th})
\end{align}

In contrast to SPSN, the PSU model omits the Bernoulli process and employs a deterministic approach to spiking, which does not impede the computation of potential events in $I=WX$. Spike estimation relies exclusively on the membrane potential $S' = AI$, whereas the actual spike discharge during the reset phase is represented by $AI-BS$, with $S$ denoting the actual spike activity. Given that the interaction between $BS$ involves matrix multiplication, the spike discharge rate at subsequent time steps tends to be elevated. Nevertheless, utilizing $S'$ in the \textit{Reset} computation reduces the firing rate compared to the traditional LIF model, offering benefits for sparse computation within SNNs.
Besides setting the hyperparameter $\tau$ similar to the LIF model, the PSU does not involve any additional hyperparameters. Its computation process exhibits higher sparsity, and the employment of parallel computation markedly enhances training and inference efficiency.

\subsection{Input-aware and Reset-aware PSU}

Besides the accelerated computational performance offered by parallel computing, another benefit of PSU in computing spike outputs through matrices is its improved proficiency in managing temporal data dynamics. In data-driven LIF models, PLIF parameterizes the leakage factor $\tau$ to balance learning new information and forgetting the old. In contrast, PSU facilitates an efficient approach to modulating information at varying moments. A practical method involves adapting the matrix $A$ based on the data. As shown in Fig. \ref{method}, the parameter matrix $A$ contributes to both the  \textit{Leak-integrate} and  \textit{Reset} process, with its primary function being the integration of diverse input information. To augment the learning efficiency, we introduce an input-aware PSU (IPSU) and a reset-aware PSU (RPSU). This approach entails dividing the leakage matrix $A$ into input and reset segments, concentrating on learning just one of these aspects.
\begin{align}
    S = \Theta(A_II - B\sigma(A_R I) - V_{th})
\end{align}

The Sigmoid function in the  \textit{Reset} process differentiates the functionalities of IPSU and RPSU. Despite sharing a common underlying purpose, they exhibit unique computational properties. RPSU is tailored to enhance information interaction between moments using a learnable parameter. Incorporating the Sigmoid function ensures that parameter alterations have a minimal effect on learning stability, rendering RPSU particularly effective for smaller datasets. In contrast, IPSU is specifically engineered to directly and efficiently learn variations between moments over prolonged durations, excelling in long-term dependency learning and consequently more likely to achieve superior outcomes.

Meanwhile, we incorporate causal masking into the learnable matrices to preclude the model from accessing future data. This necessitates that the learnable matrices consistently adhere to the form of a lower triangular matrix. In essence, this means inputs from future time steps should not influence the integration of current information. This approach ensures that the model's predictions are solely based on historical data, thereby maintaining the temporal integrity and causality within the learning process.
\begin{align}
    A_{ij} = \begin{cases}a_{ij}, \quad i \geq j \\ 0, \quad i < j \end{cases}
\end{align}
Here, $A_{ij}$ denotes the element of matrix $A$ located at the $i$-th row and $j$-th column, and $a_{ij}$ represents a specific element of matrix $A$. While this introduces additional parameters, in the context of the entire network, the incremental parameters for the learning matrix amount to approximately $(T^2+T)/2$. This increase is not significantly burdensome for network training, particularly as $T$ is usually small in direct training scenarios, in contrast to conversion-based SNNs. Furthermore, compared to conventional SNNs, PSU and its variations introduce only the estimated spike state $S'$, which is considered negligible considering the substantial simulation acceleration they enable. Implementing causal masking allows for a transition of PSU and its variants to a serial computing model during the inference phase. This means calculating $V[t]$ and $R[t]$ at each time step, reducing storage requirements.

\section{Experiments}

\subsection{Experimental Setup}

\begin{figure}[t]
    \centering
    \includegraphics[scale=0.36]{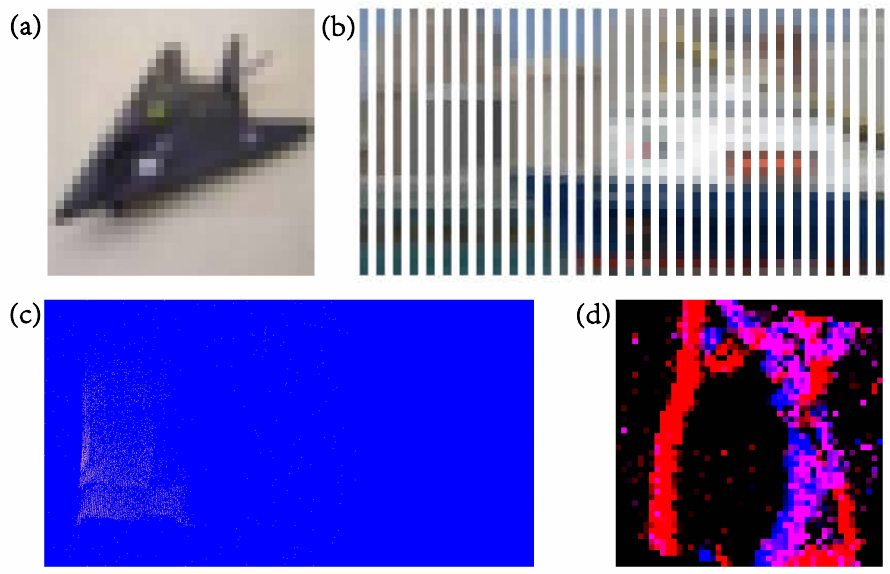}
    \caption{Example of Datasets. (a) CIFAR10 (b) Sequential CIFAR10 (c) SHD (d) DVSCIFAR10}
    \label{example}
\end{figure}

We employ the full SNN computational framework BrainCog \cite{zeng2023braincog} to validate the efficacy of the proposed PSU. Extensive experiments are conducted on four types of datasets, including static (CIFAR10) and sequential images (Sequential CIFAR10 and CIFAR100), DVS (DVSCIFAR10), and speech dataset (SHD), as shown in Fig. \ref{example}.

\begin{table}[t]
    \centering
    \caption{Experimental Setup Details.}
    \resizebox{\columnwidth}{!}{    
    \begin{tabular}{c|cccccccccccc}
        \toprule
        Dataset  & Network  &  Batch Size   &  Loss & Surrogate Function & Optimizer\\
        \midrule
        Static  &  CIFARNet \cite{guo2022recdis}, ResNet18 & 128  & CE & Atan & AdamW \\
        Sequential & 6*128C3-256-10/100 & 128  & CE & Atan & AdamW \\
        DVS  &  VGGSNN \cite{deng2022temporal} & 128 & CE & PWL & AdamW \\
        Speech & 700-1024-1024-20 & 256 & MSE & Atan & SGD \\
        \bottomrule
    \end{tabular}}
    \label{expparam}
\end{table}

\begin{figure*}[t]
    \centering
    \subfigure[Simulation Time on FC Net]{
        \label{cap1}
        \includegraphics[scale=0.2]{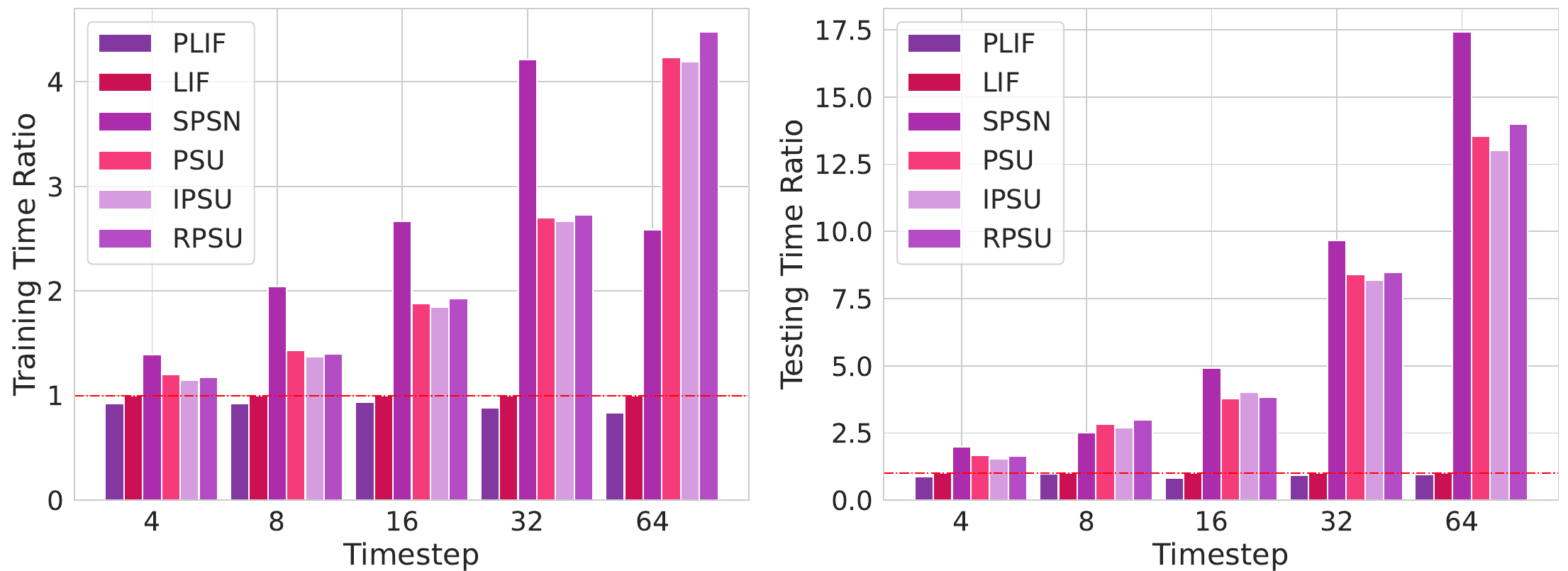}}
    \subfigure[Simulation Time on VGGSNN. ]{
        \label{cap2}
        \includegraphics[scale=0.2]{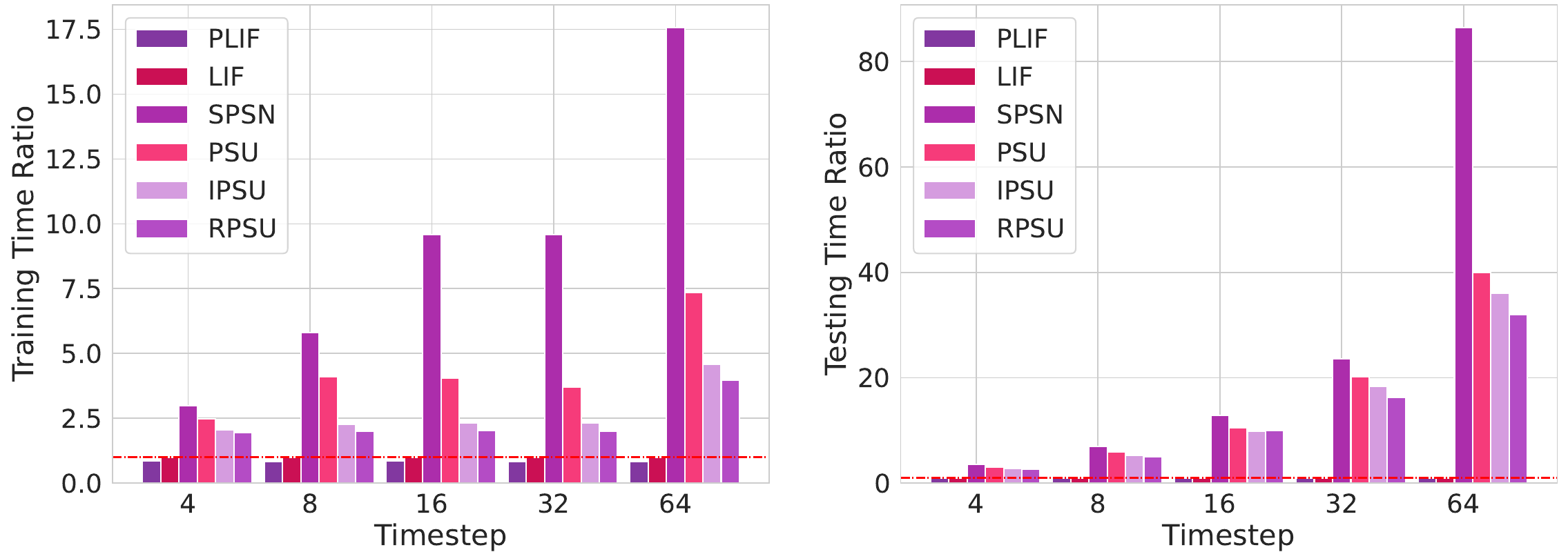}}
    \caption{The Simulation Time Ratio of the Training and Testing Phase on SHD and DVSCIFAR10 datasets.}
    \label{simutime}
\end{figure*}

\begin{figure*}[t]
    \centering
    \subfigure[Firing rate of the VGGSNN on DVSCIFAR10.]{
        \includegraphics[scale=0.22]{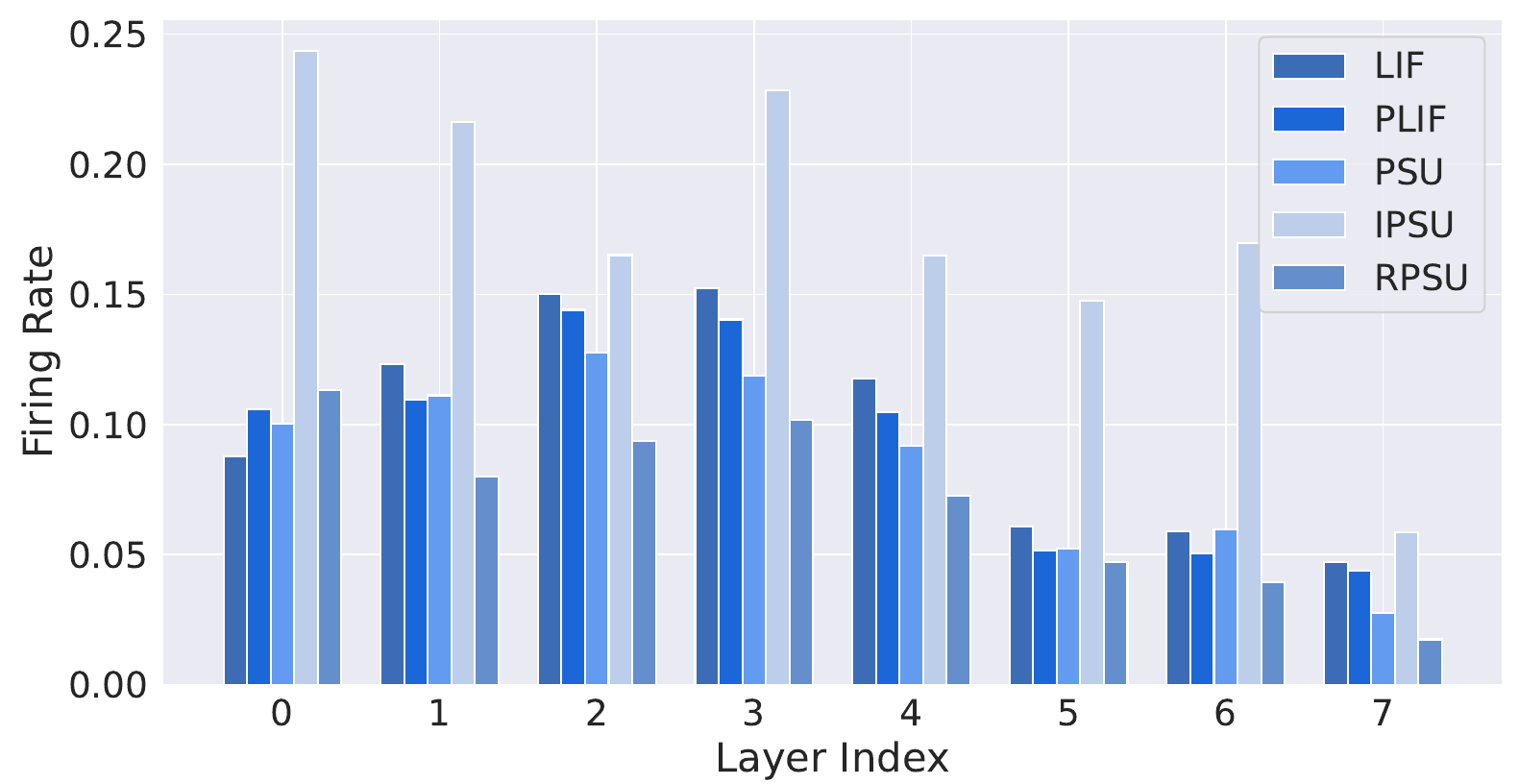}
        \label{vis1}}
    \subfigure[The feature map of the first and third layers of the VGGSNN on DVSCIFAR10.]{
        \includegraphics[scale=0.22]{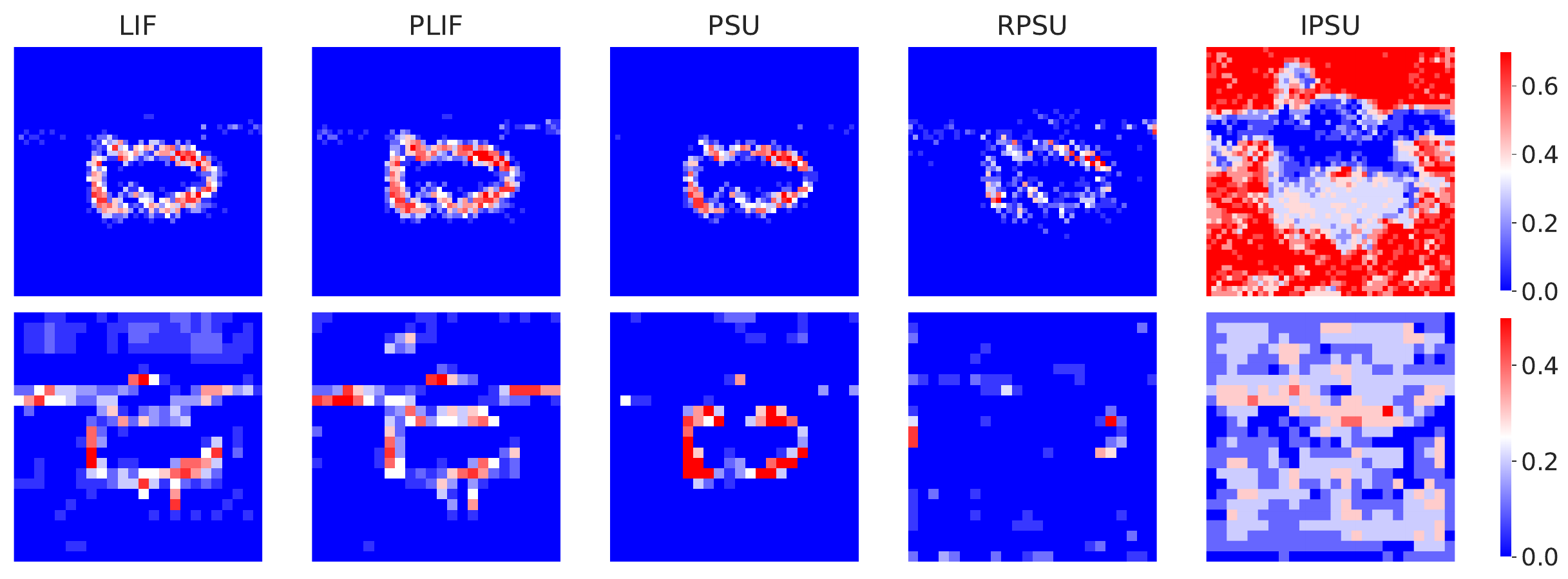}
        \label{vis2}}
    \caption{Analysis of neural activity in VGGSNN on DVSCIFAR10, including (a) firing rates and (b) feature maps.}
    \label{fig:enter-label}
\end{figure*}


We utilize various model configurations to accommodate different data inputs and ensure fair comparisons. CIFARNet, a class VGG model akin to that in \cite{guo2022recdis}, is employed for static image datasets. For the Sequential CIFAR dataset, a 6-layer one-dimensional convolutional structure complemented by two fully connected layers is used for output decoding. In the case of the DVS datasets, we follow the approach of \cite{deng2022temporal}, using a class VGG model, specifically VGGSNN. A four-layer, fully connected network is implemented to classify the SHD dataset. Direct encoding is used to enhance training efficiency, and the output neuron threshold is set to infinity.

To optimize the training of SNNs and improve the model's generalization capabilities, we implement the arctangent (Atan) \cite{fang2021incorporating} and Piece-wise linear (PWL) \cite{neftci2019surrogate} function as the surrogate gradient function. This addresses the challenge posed by the non-differentiable nature of spikes in training with backpropagation. Additionally, we adopt a cosine annealing strategy for learning rate adjustment and incorporated a label smoothing technique to enhance performance. The additional parameters relevant to our methodology are outlined in Tab. \ref{expparam}. CE and MSE stand for Cross-Entropy Loss and Mean Squared Error Loss. And 
AdamW and SGD refer to the AdamW optimizer and the Stochastic Gradient Descent optimizer.

\subsection{Simulation Time Saving}

The primary aim of this study is to enhance simulation speed by parallelizing the dynamic process of spiking neurons, encompassing training and testing phases. Fig.~\ref{simutime} illustrates the comparative simulation speeds between our proposed parallel spiking neuron model and the traditional LIF model, PLIF, and SPSN across SHD and DVS datasets. We examine five different timestep settings: 4, 8, 16, 32, and 64, measuring the time required for ten epochs on the full dataset to derive the mean value. The vertical axis in the graph represents the speed ratio of the current neuron model to the LIF neuron of identical configuration during training and testing. It's noteworthy that while SPSN demonstrates considerably faster simulation speeds, its training stability is questionable, particularly in datasets like DVSCIFAR10, where its test accuracy is merely 12\%, as shown in Fig. \ref{cap2}. Consequently, SPSN is excluded from further experiments. Our results indicate a substantial improvement in simulation speed compared to the native LIF model, which is expected to significantly enhance the efficiency of SNNs in optimization and deployment. Despite IPSU and RPSU incorporating learnable parameters, their inference times during training and testing are approximately equal. In the graph, the time ratio during training is lower than that in pure inference for the same timestep, as the backpropagation phase constitutes a larger fraction of the total computation time. Notably, in the inference phase, the PSU achieves a greater VGGSNN acceleration than FCNet, attributable to the former's deeper layer structure, which benefits more from full parallel computation.

\begin{table*}[t]
	\centering
    \caption{Comparison with previous work on different datasets.}
    \begin{tabular}{c|cccccccccccc}
        \toprule
        Dataset  & Method  & Parallel & Architecture  & Time Step  & Accuracy(\%)\\
        \midrule
        \multirow{8}{*}{CIFAR10} 
        & PLIF \cite{fang2021incorporating} & \XSolidBrush & ConvNet & 4 & 93.50 \\
        & TET \cite{deng2022temporal} & \XSolidBrush & ResNet19 & 4 & 94.44 \\
        & SLTT \cite{meng2023towards} & \XSolidBrush & ResNet18 & 6 & 94.40 \\
        & ResDis-SNN \cite{guo2022recdis} & \XSolidBrush & CIFARNet & 4 & 92.20 \\
        & ASGL \cite{wang2023adaptive} & \XSolidBrush & CIFARNet & 4 & 94.74 \\
        & PSN \cite{fang2023parallel} & \CheckmarkBold & Modified PLIF Net & 4 & 95.32 \\
        & PSU, IPSU, RPSU & \CheckmarkBold & CIFARNet & 4 & 94.80, \textbf{95.48}, 94.78 \\ 

        \midrule
        \multirow{2}{*}{Sequential CIFAR10} 
        & PLIF \cite{fang2021incorporating} & \XSolidBrush & ConvNet & 32 & 81.47 \\
        & PSU, IPSU, RPSU & \CheckmarkBold & ConvNet & 32 & 79.52, \textbf{87.28}, 84.44 \\ 

        \midrule
        \multirow{2}{*}{Sequential CIFAR100} 
        & PLIF \cite{fang2021incorporating} & \XSolidBrush & ConvNet & 32 & 53.38 \\
        & PSU, IPSU, RPSU & \CheckmarkBold & ConvNet & 32 & 51.65, \textbf{59.76}, 57.89 \\ 
        
        \midrule
        \multirow{7}{*}{DVSCIFAR10} 
        & PLIF \cite{fang2021incorporating} & \XSolidBrush & ConvNet & 10 & 74.80 \\
        & Dspike \cite{li2021differentiable} & \XSolidBrush & ResNet18 & 10 & 75.40 \\
        & TET \cite{deng2022temporal} & \XSolidBrush & VGGSNN & 10 & 77.40 \\
        & ASGL \cite{wang2023adaptive} & \XSolidBrush & VGGSNN & 10 & 78.90 \\
        & SLTT \cite{meng2023towards} & \XSolidBrush & VGG-11 & 10 &  \textbf{82.20} \\
        & ResDis-SNN \cite{guo2022recdis} & \XSolidBrush & CIFARNet & - & 67.30 \\
        & PSU, IPSU, RPSU & \CheckmarkBold & VGGSNN & 10 & 81.40, 75.30, 82.00 \\

        \midrule
        \multirow{5}{*}{SHD} 
        & Heterogeneous RSNN \cite{perez2021neural} & \XSolidBrush & - & - & 82.70 \\
        & ASGL \cite{wang2023adaptive} & \XSolidBrush & FCNet & 15 & 86.90 \\
        & TA-SNN \cite{yao2021temporal} & \XSolidBrush & FCNet & 15 & 91.08 \\
        & SPSN \cite{yarga2023accelerating} & \CheckmarkBold & FCNet & 15 & 89.73 \\
        & PSU, IPSU, RPSU & \CheckmarkBold & FCNet & 15 & 90.64. 91.30, \textbf{92.49} \\
        \bottomrule
    \end{tabular}
    \label{all}
\end{table*}

\subsection{Neural Activity Analysis}

Beyond the speed advantage in simulation, we expect that the computational processes in spiking neurons will demonstrate greater sparsity, potentially further diminishing the energy consumption required for computations in the SNN upon deployment. Our analysis of the spike issuance rate within the initial eight layers of VGGSNN during the classification of DVSCIFAR10, as illustrated in Fig. \ref{vis1}, supports this anticipation. The data suggests that the PSU architecture can conduct computations with reduced spike activity. In contrast, the IPSU model, which utilizes learnable parameters to adjust the membrane potential $V$, shows increased spike activity following the learning phase. On the other hand, RPSUs, which integrate learnable parameters within the Sigmoid function, exhibit more consistent learning outcomes and enhanced handling of dynamic information interactions while maintaining a lower spike activity level than that observed in the LIF model. Moreover, despite its rapid simulation capabilities, the SPSN model demonstrates instability in neural activity and proves inefficient training due to the lack of a reset mechanism. Similarly, the PLIF model, which actively and adaptively adjusts leakage coefficients in response to incoming data, also does not achieve the level of sparsity seen in the PSU model.
In light of the observation that PSU and RPSU can maintain comparable or superior performance with less frequent spiking activity, we delved deeper into this phenomenon. We apply these models to identical data samples and visualized neuron activities in the first and third layers, as depicted in Fig. \ref{vis2}. The findings highlight that PSU and its derivatives are adept at minimizing non-essential features' representation while preserving pivotal elements. This starkly contrasts the IPSU model, which diverges from the LIF model in its learning approach, leading to a unique pattern in spiking activity.

\begin{figure}[t]
    \centering
    \includegraphics[scale=0.22]{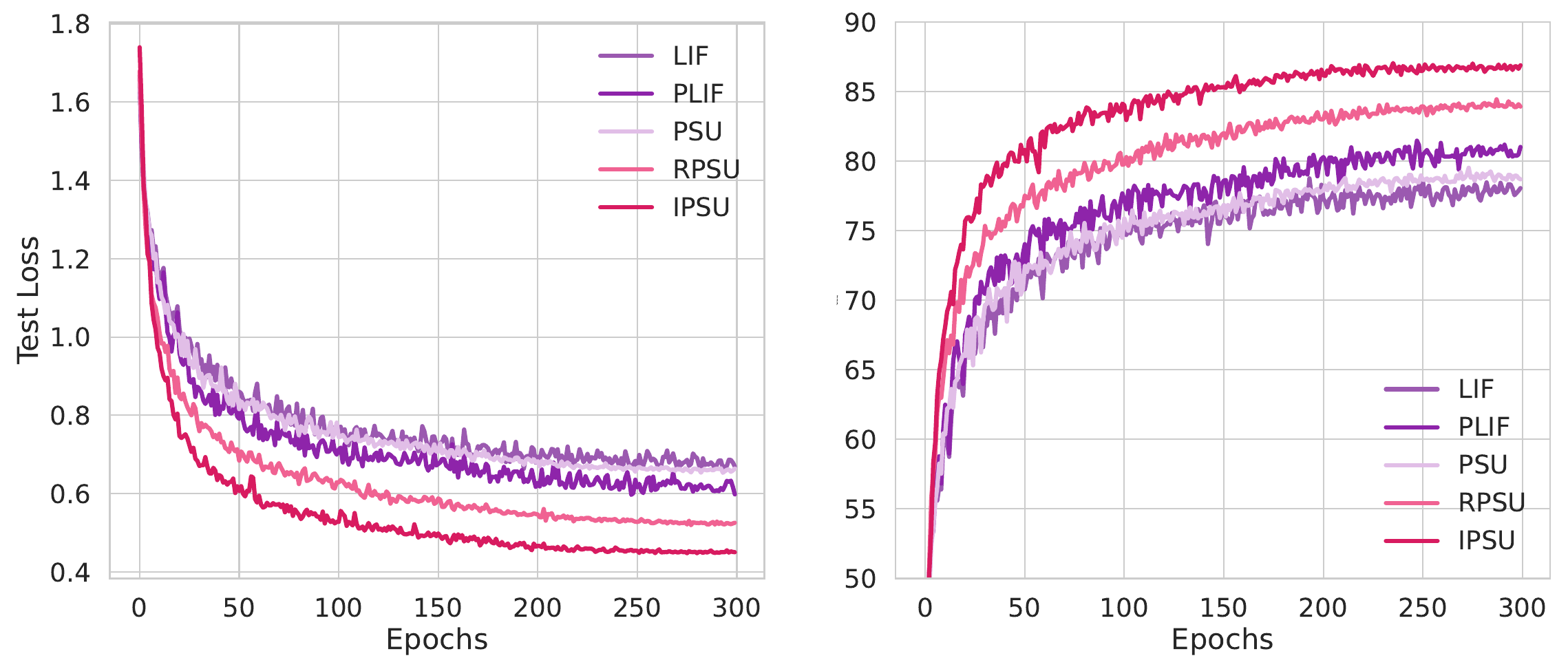}
    \caption{The test accuracy and loss curves of different models' performance on the Sequential CIFAR10.}
    \label{secifartest}
\end{figure}

\subsection{Comparison with Previous Work}

In this section, we compare our methodology with various state-of-the-art approaches, including the LIF model, PLIF \cite{fang2021incorporating}, SPSN \cite{yarga2023accelerating}, Dspike \cite{li2021differentiable}, TET \cite{deng2022temporal}, ASGL \cite{wang2023adaptive}, SLTT \cite{meng2023towards}, and TA-SNN \cite{yao2021temporal}. This comparison aimed to validate our method's effectiveness across diverse datasets, encompassing static images, sequential images, DVS data, and speech datasets.
The results, as presented in Tab. \ref{all}, clearly illustrate our method's superiority in performance across each dataset type. Specifically, in static image datasets, our approach not only achieved a 0.38\% accuracy advantage over the efficient training method SLTT, requiring fewer time steps but also surpassed the accuracy metrics reported by existing parallel neuron works such as PSN. In sequential datasets like Sequential CIFAR10, we observed that our PSU model outperforms the traditional LIF model regarding simulation speed, demonstrating superior performance advantages. The test curves for different models in this dataset category are elaborately depicted in Fig. \ref{secifartest}.
Models with learnable parameters generally exhibited enhanced performance compared to LIF and standard PSU models. In the Sequential CIFAR10 dataset, for instance, IPSU exceeds the performance of both PSU and RPSU, marking an impressive accuracy of 87.28\%. However, PLIF, despite its incorporation of learnable parameters, is limited to an accuracy of 81.47\% within the serial optimization framework. Interestingly, in datasets such as DVSCIFAR10 and SHD, which are comparatively smaller, IPSU did not perform as well as RPSU. This is attributed to IPSU's higher tendency towards overfitting, whereas RPSU's parameter changes do not significantly impact the stability of the learning process.
In datasets abundant with temporal information, like the SHD datasets, both IPSU and RPSU demonstrate exceptional performance, eclipsing existing methodologies. Specifically, RPSU achieved a significant 2.76\% performance improvement relative to SPSN, highlighting its effectiveness in handling rich temporal data. 
Conclusively, the experimental results indicate that PSU and its variants not only match but often surpass the performance of other methods in all evaluated datasets. This strongly suggests that our approach successfully balances simulation speed, computational sparsity, and accuracy.

\section{Conclusion}

The acceleration of spiking units is a topic of interest but directly determines the efficiency of SNN applications. This paper presents a significant advancement in SNNs by developing Parallel Spiking Units and their variants, IPSU and RPSU. By innovatively decoupling the leaky integrate-and-fire mechanism and introducing a probabilistic reset process, this approach enables parallel computation of neuron membrane potentials, significantly enhancing computational efficiency and speed. Our findings demonstrate that PSU outperforms traditional LIF neurons and existing spiking neuron models in various datasets, offering sparser spiking activity and faster processing. Importantly, this study addresses the critical limitation of SNNs in utilizing parallel computing resources like GPUs, paving the way for low-power AI systems.

\section*{Acknowledgment}
This research was financially supported by Postdoctoral Fellowship Program of CPSF (Grant No. GZC20232994) and a funding from Institute of Automation, Chinese Academy of Sciences (Grant No. E411230101).

\bibliographystyle{IEEEtran}
\bibliography{mybib}

\end{document}